\documentclass[10pt,twocolumn]{article}
\usepackage[preprint]{antintlpaper}
\usepackage[utf8]{inputenc}
\usepackage{inconsolata}
\newcommand{\code}{\textsc{Code}}
\newcommand{\wm}{\textsc{WM}}
\newcommand{\na}{\textit{n/a}}
\newcommand{\stitle}[1]{\vspace{1mm} \noindent {\bf #1}}

\AntTitle{Which World-Model Errors Matter for Training Language Agents?}
\AntAuthors{Yuchen~Mao\AntEqualContributor \and
  Zhihao~Wen\AntEqualContributor \and
  Yiwen~Zhao\AntEqualContributor \and
  Sijie~Dong\texorpdfstring{\textsuperscript{\dag}}{} \and
  Yimin~Wang \and
  Pan~Wang \texorpdfstring{\\}{, }
  Hang~Wang \and
  Yitong~Lin \and
  Yihao~Hu \and
  Jiang-Ming~Yang \and
  Wei~Wu\texorpdfstring{\textsuperscript{\dag}}{}}
\AntAffiliations{Ant International, Ant Group}
\AntFirstPageNote{\textsuperscript{*}\,Equal Contributions\quad
  \textsuperscript{\dag}\,Corresponding Authors}
\AntDate{November 17, 2026}
\AntAbstract{%
Language agents can practice in text world models instead of slow real environments, but simulated feedback is often wrong, and checking it by execution gives up the savings. Which errors are worth checking? Not necessarily the likeliest mislabels. Group-relative learners such as GRPO score each attempt against its peers, so errors differ sharply in how much they change the learning signal. In WebShop, executable and simulated feedback prove complementary. After a short executable warm-up, mixing them ends above either alone, while executing 40\% fewer groups than executable training. Replay shows that simulator errors are common, nearly always distort advantages, and are mostly false successes. A group's reward structure, more than predictor accuracy, decides which checks change advantages most. These findings yield advantage-change querying, which spends scarce execution on the groups whose advantages a CPU classifier expects to change most. Continuing a world-model-trained policy with a quarter of groups checked, it gains 7.5 points of strict success over random checks. It also outperforms heuristic and error-based checks, improves other learners, and its selections transfer to frozen ALFWorld rollouts. World models should therefore be judged by which errors reach the learner, not by label accuracy alone.
}

\begin{document}
\raggedbottom
\makeanttitle

\section{Introduction}
Training language agents with reinforcement learning requires many interactions. In real environments these interactions can be slow, rate-limited, or costly, because pages must load, tools must run, and some actions, such as purchases, cannot be undone. Text world models reduce this cost by predicting the next observation and the outcome, so an agent can practice on simulated web pages and tool outputs \citep{chae2025web,gu2025webdreamer,li2026word}. They also cheaply supply the many attempts per task that group-relative methods need, which makes simulator training attractive \mbox{\citep{yang2026wmrl}}.

\begin{figure}[t]
\centering
\includegraphics[width=.85\linewidth]{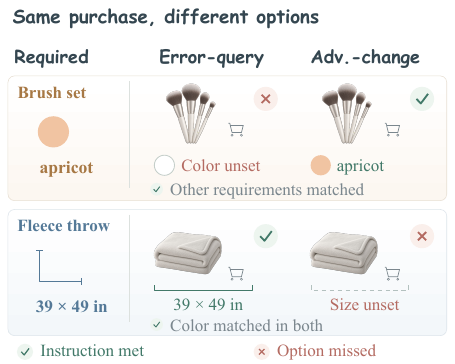}
\caption{Strict success requires every requested option.} 
\label{fig:cases}
\end{figure}

The price is that the simulator is sometimes wrong, and a learner may exploit its mistakes. In WebShop \citep{yao2022webshop}, a task is \emph{strict success} only if every requested option is satisfied (Figure~\ref{fig:cases}). A policy trained for 120 updates in a public world model (\wm{}) reaches 51.5\%, compared with 65.0\% in the executable environment (\code{}). Executable interaction corrects the feedback but spends the calls simulation was meant to save.

Spending execution well raises three challenges. \textbf{C1:~Combining two imperfect sources.} Executable feedback is accurate but slow, and simulated feedback is cheap but wrong. It is unclear when execution pays off in training, and whether its benefit comes from correct observations or from correct rewards. \textbf{C2:~Error rates hide which errors matter.} Group Relative Policy Optimization (GRPO; \citealp{shao2024deepseekmath}) uses a label only through an advantage that scores the attempt against the others at the same task. One wrong label can therefore move every advantage in its group, while labels that are all wrong in the same way move none. \textbf{C3:~Choosing checks before seeing outcomes.} When execution is scarce, a rule must pick groups to check from simulated evidence alone, at a cost far below the execution it saves. The natural choice is to check the likeliest mislabels, which targets error rate, not the advantage change that C2 exposes.

For C1 and C2, we measure what executable feedback changes in a group-relative learner (Section~\ref{sec:measurement}). The two sources are complementary: after a 30-update executable warm-up, mixed schedules end above either source alone while training fewer groups in \code{}, and executable rewards help even when observations stay simulated. Correcting simulated labels changes the learner's advantages in nearly every group with an error. Most of these errors are false successes, simulated purchases that fail in execution, and their count more than triples during simulator training. A group's reward structure, more than predictor accuracy, identifies these changes. For C3, \emph{advantage-change querying} (Section~\ref{sec:method}) therefore checks the groups whose advantages are expected to change most, using a frozen TF-IDF classifier that runs on one CPU thread.
Our contributions are threefold:
\begin{itemize}[leftmargin=*,topsep=0pt,itemsep=1.5pt,parsep=0pt,partopsep=0pt]
    \item We formulate \emph{advantage error}, the change a simulator error causes in a group-relative learner's advantages, and measure when and where executable feedback matters (C1, C2; Section~\ref{sec:measurement}).
    \item We develop \emph{advantage-change querying}, a check rule that spends scarce execution on groups whose advantages are expected to change most (C3; Section~\ref{sec:method}). It scores advantage change and uses one frozen CPU classifier and keeps all unchecked groups.
    \item We show that this rule beats random, heuristic, and error-based checks, gaining 7.5 points of strict success over random checks after world-model training; it also improves other learners, and its selections transfer to frozen ALFWorld rollouts (Sections~\ref{sec:results} and~\ref{sec:transfer}).
\end{itemize}

\section{Related Work}
\label{sec:related}

\begin{table*}[t]
\centering\small\setlength{\tabcolsep}{8pt}\renewcommand{\arraystretch}{1.12}
\begin{tabular}{llrrrrr}
\toprule
Feedback schedule & Shared prefix & 30 & 60 & 90 & 120 & 150\\\midrule
Pure \code{} & None & 37.0 & 52.0 & 66.0 & 65.0 & 63.5\\
Pure \wm{} & None & 29.5 & 48.5 & 41.5 & 51.5 & 56.5\\
Random 50\% full Code interaction & None & 33.0 & 47.0 & 53.0 & 59.0 & 61.0\\\midrule
Code followed by WM & Code75 & \multicolumn{2}{c}{shared Code prefix} & 56.5 & 57.0 & 58.5\\
WM followed by Code & WM75 & \multicolumn{2}{c}{shared WM prefix} & 59.5 & 57.5 & 62.5\\\midrule
Decreasing Code fraction & Code30 & 37.0 & 57.5 & 57.5 & 65.5 & 64.5\\
Increasing Code fraction & Code30 & 37.0 & 44.5 & 62.5 & 63.0 & 67.0\\
Random 50\% after the prefix & Code30 & 37.0 & 55.5 & 59.5 & 52.5 & \textbf{68.5}\\
\bottomrule
\end{tabular}
\caption{Feedback schedules: strict success (\%) on the 200 public test tasks (Primary200) after each number of updates. Full \code{} interaction supplies observations and rewards. A prefix names the source and length of a shared initial phase (e.g., Code30 and WM75 denote 30 updates in \code{} and 75 in \wm{}). Bold marks the best endpoint. Appendix~\ref{app:schedules} gives the group counts.}
\label{tab:schedulesmain}
\label{tab:trajectories}
\end{table*}

\stitle{Agents and world models.}
Language agents act in web, embodied, and text-game environments \citep{yao2022webshop,shridhar2021alfworld,wang2022scienceworld,cote2018textworld} via prompting \citep{yao2023react} or multi-turn RL \citep{zhou2024archer,qi2025webrl,wang2025ragen,zhao2026faca}, including GRPO \citep{shao2024deepseekmath,guo2025deepseekr1} and its variants \citep{yu2025dapo,liu2025understanding,feng2025gigpo,lu2026sdar,li2026tepo}. Learning from simulated experience dates to Dyna \citep{sutton1991dyna,hafner2020dream}; model-based RL \citep{abbeel2006using,talvitie2017self,janner2019mbpo,ross2012agnostic} and sim-to-real transfer \citep{tobin2017domain,hanna2017grounded} address imperfect simulators. LLM world models now support planning \citep{hao2023rap,chae2025web,gu2025webdreamer} and agent training \citep{ding2026dynaweb,li2026word,wang2026awm}, despite unreliable state predictions \citep{wang2024textsim}, and world-model RL (WMRL; \citealp{yang2026wmrl}) calibrates simulated rewards with executable checks. We instead measure which simulator errors reach the learner, compare feedback schedules, and choose which groups to check when execution is scarce.

\stitle{Noisy and decision-aware feedback.}
Prior work studies corrupted, flipped, and spurious rewards, including in GRPO \citep{everitt2017reinforcement,wang2020perturbed,elmansouri2025noise,wang2026spurious}, and reward hacking \citep{skalse2022defining,pan2022effects,gao2023scaling,coste2024reward}. Decision-aware model learning \citep{farahmand2017vaml,farahmand2018iterative,ayoub2020model,abachi2020policy,lambert2020objective,grimm2020value,lovatto2020decision} and decision-focused learning \citep{donti2017task,wilder2019melding,elmachtoub2022smart} judge models by the decisions they support, and reward-model accuracy \citep{lambert2025rewardbench} need not predict policy quality \citep{chen2024accuracy,razin2025teacher}. We likewise judge world-model errors by the learner advantages they change.

\stitle{Label acquisition.}
Active learning queries labels by uncertainty, expected error reduction or model change, or likely verifier errors \citep{cohn1996active,roy2001toward,settles2007multiple,kapoor2007selective,freytag2014selecting,konyushkova2017learning,kirsch2019batchbald,ash2020badge,lightman2024lets}, though its probabilities can be miscalibrated \citep{guo2017calibration,malik2019calibrated,ovadia2019can} and its samples biased \citep{farquhar2021statistical}. In RL, agents can pay to observe rewards \citep{krueger2020active}, query policy-relevant preferences \citep{christiano2017deep,lee2021pebble,lindner2021idrl}, or prioritize replay \citep{schaul2016prioritized}. CARE \citep{wang2026care} also queries by expected advantage change in RL with verifiable rewards, using online hidden-state classifiers and dropping unreliable unqueried samples. We instead train multi-turn agents in a world model and keep every unchecked group, since most simulated labels are correct (Section~\ref{sec:measurement}).

\section{When and Where Does Executable Feedback Matter?}
\label{sec:measurement}
\stitle{Setting.}
We study WebShop with the public text world model \texttt{WorldModel-Webshop-Qwen2.5-7B} as the simulator (\wm{}) and the executable benchmark (\code{}) as ground truth \citep{li2026word}. Each task asks the agent to search a product catalog and buy an item with the requested attributes, options, and price. A Qwen2.5-7B-Instruct agent \citep{qwen2025qwen25} is trained with GRPO, which samples eight attempts per task and compares each attempt with the others. Strict success means a purchase that meets every requirement and earns WebShop's full reward (Figure~\ref{fig:cases}). In \wm{}, the world model also predicts whether the episode succeeds, and this prediction serves as the reward. We score policies on the 200 public test tasks (Primary200).

\subsection{Finding 1: Mixed feedback outperforms either source alone}
Table~\ref{tab:schedulesmain} compares schedules in which each task group is trained either with full \code{} interaction or in \wm{}. Neither source alone is best. At update 150, pure \code{} training reaches 63.5\% and pure \wm{} training 56.5\%. After 30 \code{} updates, all three mixed schedules end higher, at \mbox{64.5--68.5\%}, and each trains the same 1,440 groups in \code{}, against 2,400 for pure \code{}. The schedule that gradually increases the \code{} fraction climbs steadily to 67.0\%, and routing a random half of the groups to \code{} reaches 68.5\% after a dip at update 120. Both exceed the best pure-\code{} checkpoint (66.0\% at update 90), which has trained the same 1,440 groups in \code{}. Without the warm-up, the same random mixture ends at 61.0\% with 1,200 \code{} groups, and without later mixing, switching entirely to \wm{} after 75 \code{} updates ends at 58.5\%. In these runs, the best schedules start in \code{} and keep mixing.

\begin{table}[t]
\centering\small\setlength{\tabcolsep}{4pt}
\begin{tabular}{lr}
\toprule
Reward intervention & Success (\%)\\\midrule
\multicolumn{2}{l}{\textit{(a) Same update-150 parent; evaluate at 180}}\\
Keep WM rewards & 51.0\\
Replace with Code rewards & \textbf{60.0}\\\midrule
\multicolumn{2}{l}{\textit{(b) Same WM75 parent; evaluate at 150}}\\
No correction (history treatment only) & 45.0\\
Correct false successes & \textbf{60.5}\\
Correct false failures & 52.5\\
Correct both error types & 55.0\\
\bottomrule\end{tabular}
\caption{Isolating the reward channel: Primary200 strict success (\%). Within each block, runs share the parent checkpoint and history treatment and differ only in rewards. In (b), corrections apply to a random half of the groups in each update. Bold: best per block.}
\label{tab:attributionmain}
\end{table}

\subsection{Finding 2: Executable rewards alone change learning}
Full interaction changes observations and rewards together. To isolate the rewards, we continue an update-150 checkpoint twice, changing only the reward source (Table~\ref{tab:attributionmain}a). This parent checkpoint was trained with a \emph{history treatment}: each action also runs in \code{}, and the world model predicts the next page from this \code{} history instead of its own earlier predictions. The agent still sees simulated pages (\mbox{Appendix}~\ref{app:channels}). Both continuations keep this treatment. After 30 more updates, the continuation with executable rewards reaches 60.0\%, against 51.0\% with simulated rewards. We also continue WM75, the checkpoint after 75 \wm{} updates, and give a random half of the groups executable feedback (Appendix Table~\ref{tab:channels}). At update 150, the history treatment alone reaches 45.0\%, \mbox{below} pure \wm{} training (56.5\%). Adding executable rewards to it reaches 55.0\%, executable rewards alone reach 54.0\%, and full interaction reaches 63.5\%. Rewards alone lead pure \wm{} training at updates 90 and 120, but not at 150. With the history treatment, fixing only the false successes reaches 60.5\%, above the 56.5\% of pure \wm{} training at the same update (Finding~4).

\begin{figure}[t]
\centering
\includegraphics[width=\linewidth]{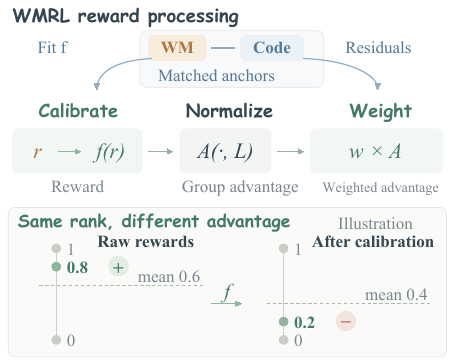}
\caption{Group-relative advantages depend on how rewards are spaced, not only on their order. A monotone map $f$ turns rewards $[0,.8,1]$ into $[0,.2,1]$, preserving their order but flipping the middle episode's advantage from positive to negative. Top: WMRL (Section~\ref{sec:conditions}) fits this map to checked \wm{}/\code{} reward pairs, calibrates unchecked rewards before normalization, and weights checked groups' advantages by the pairs' residuals.}
\label{fig:creditmechanism}
\end{figure}

\subsection{Finding 3: Simulator errors are common and nearly always change advantages}
A \emph{check}, unlike these continuations, rewrites no history. It replays an episode's recorded actions in \code{} and replaces its simulated label with the executable one, keeping all simulated observations. GRPO uses the binary label only through the attempt's \emph{task advantage} $A_i=(r_i-\mu)/\sigma$, where $\mu$ and $\sigma$ are the group's reward mean and standard deviation, weighted by action counts, and $A_i=0$ when all rewards are equal. Format penalties are left out (Appendix~\ref{app:credit} gives the exact form). The task advantage sets whether and how strongly the attempt is encouraged relative to its peers. Because it depends on the group's mean and spread, changing one reward can move every advantage in the group. Even an order-preserving change of rewards can flip an advantage's sign (Figure~\ref{fig:creditmechanism}). A group's \emph{advantage error} is the mean absolute change in its task advantages when all its labels are checked.

To audit reward errors, we take nine checkpoints, three from pure world-model training and six from runs with the check rules of Section~\ref{sec:method}. We sample eight rollouts on each of 128 development tasks from each checkpoint and replay all 9,216 rollouts in \code{} (Appendix~\ref{app:measurementprotocol}). The simulator mislabels \mbox{4.4--14.0\%} of rollouts, and \mbox{22.7--44.5\%} of groups contain at least one wrong label. These errors also reach the learner's advantages. Of the 434 error-containing groups, 422 change their task advantages, with a median advantage error of .678. The 12 exceptions have all labels wrong in the same direction, such as eight simulated successes that all fail when executed. Their rewards stay identical within the group, so their advantages remain zero. The errors also grow as the policy trains inside the simulator. On the same tasks, pure world-model training produces 37 false successes per 1,024 rollouts at update 30 and 131 at update 75, while false failures stay between 8 and 12 (Table~\ref{tab:frozenlocalization}).

Label disagreement and advantage change can also come apart completely. In the tool environments of AWM \citep{wang2026awm}, a code reference and an LLM judge disagree on 1,147 of 1,152 labels, yet 139 of 144 groups have zero task advantages under both (Appendix~\ref{app:awm}).

\subsection{Finding 4: Advantage error concentrates in false successes}
Fixing only false successes removes \mbox{72.2--89.5\%} of the advantage error, and fixing only false failures removes \mbox{9.2--25.4\%}. False successes are also \mbox{2.9--10.9} times more frequent than false failures. In training with the history treatment and corrections on a random half of groups, fixing false successes helps most, reaching 60.5\% against 52.5\% for false failures and 45.0\% without reward correction (Table~\ref{tab:attributionmain}b). Correcting both reaches 55.0\%, below false successes alone. In replayed ALFWorld rollouts, however, false failures outnumber false successes (97 against 45), so the dominant type depends on the environment, and a check rule should cover both types.

\begin{figure}[t]
\centering
\includegraphics[width=\linewidth]{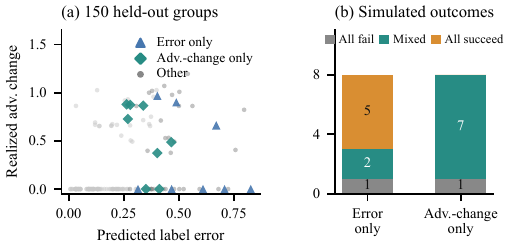}
\caption{Ranking groups by predicted error versus expected advantage change, with the same predictor at a 25\% budget on 150 held-out groups. (a) Five of the eight groups chosen only by error ranking change no advantage when checked, against two of the eight chosen only by advantage change. (b) Error ranking favors groups that the simulator marks all-success.}
\label{fig:acquisition}
\end{figure}

\subsection{Finding 5: Reward structure finds most of the captured advantage change}
A check rule must choose groups before replay, so we judge it by its \emph{captured advantage change}, the mean realized advantage change of the groups it selects. On 171 historical validation groups at a 4-of-16 budget, random checks capture .235. Two content-blind rules nearly double this. Expected advantage change (Section~\ref{sec:method}) with one historical success rate for all episodes captures .420, and the standard deviation of simulated rewards captures .446. Mixed groups tend to change most, because checking an equal-reward group often leaves its rewards equal and its advantages zero (Figure~\ref{fig:acquisition}). Learned probabilities from a TF-IDF classifier raise the capture to .471 with advantage-change scoring, against .429 for error ranking, with overlapping intervals (Appendix Table~\ref{tab:structural}). If the executable labels were known, ranking by error count would reach 94.5\% of the best possible capture at this budget. The two rules thus differ mainly in how they treat uncertain predictions (Appendix~\ref{app:prevalencecontrols}).

\begin{figure}[!t]
\centering
\includegraphics[width=\linewidth]{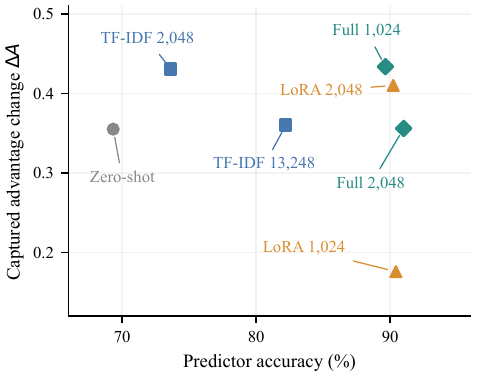}
\caption{More accurate predictors do not consistently select groups with larger advantage change. Each point is a predictor choosing 25\% of 64 held-out groups (512 trajectories) by expected advantage change. Labels give fitting-set sizes. InnerJev variants are zero-shot, LoRA-adapted, or fully fine-tuned (Appendix Table~\ref{tab:predictor}).}
\label{fig:predictorcapacity}
\end{figure}

Beyond reward structure, higher predictor accuracy does not consistently yield more captured advantage change (Figure~\ref{fig:predictorcapacity}). Across seven predictors, accuracy ranges from 69\% to 91\%, while captured advantage change (.176 to .434) does not rise with it. With the same 2,048 labels, LoRA-adapted and fully tuned versions of a 4.2B-parameter neural predictor (InnerJev; \citealp{innerjev2026}) are about 90\% accurate, against 73.6\% for a TF-IDF classifier. Yet the classifier captures more advantage change (.431) than either InnerJev version (.410 and .356). A cheap classifier is thus competitive for ranking groups.

\begin{figure*}[t]
\centering
\includegraphics[height=.567\textwidth]{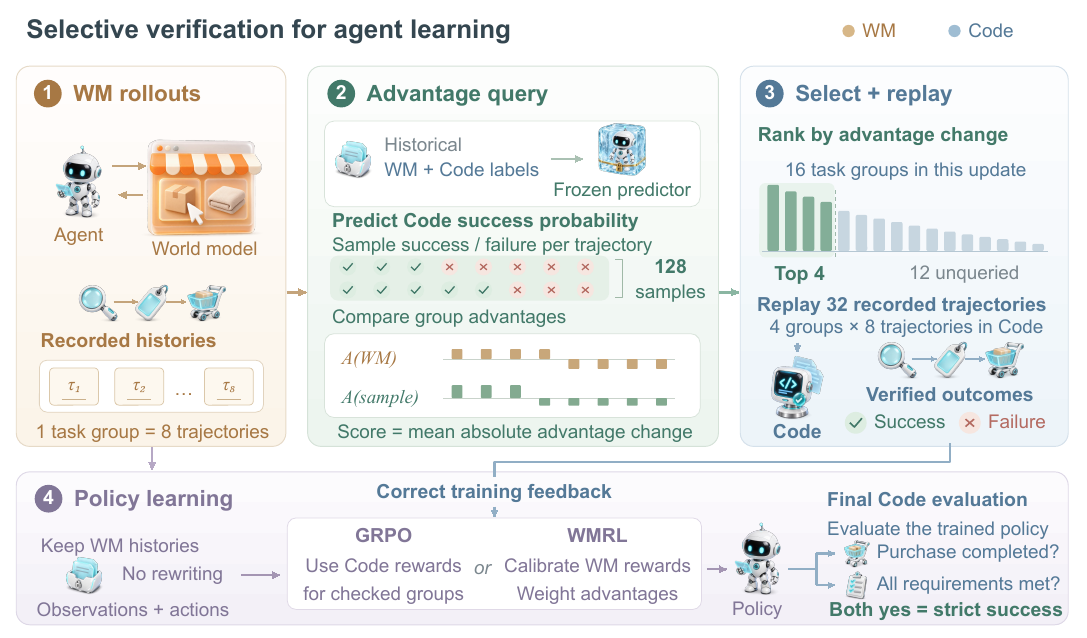}
\caption{Advantage-change querying at a 25\% budget. A frozen predictor gives each simulated episode a success probability, and hypothetical outcomes sampled from these probabilities estimate how much each group's task advantages would change. The four highest-scoring of 16 groups are then replayed in \code{}. Checked outcomes replace GRPO rewards or inform WMRL's calibration and weighting, while training observations remain simulated. All depicted outcomes and scores are illustrative. Our core code is in the supplementary software.}
\label{fig:method}
\end{figure*}

\section{Advantage-Change Querying}
\label{sec:method}
Finding~1 favors mixing the two sources when execution is affordable. When few groups can be checked, Findings~\mbox{2--5} suggest checking rewards, ranking groups by expected advantage change, and using a cheap predictor. Advantage-change querying (\emph{Adv.-change}) follows this recipe (Figure~\ref{fig:method}).

\stitle{Scoring a group by its expected advantage change.}
In GRPO, a checked reward enters the update only through the group's task advantages. Consider an all-success group whose episodes all have near-zero success probabilities. Its predicted error is high, yet a check would almost surely turn every reward into a failure, leaving the rewards equal and the advantages zero. DAPO's dynamic sampling drops such zero-advantage groups for the same reason \citep{yu2025dapo}. One likely-wrong label in a mixed group, by contrast, can move every advantage. For a group $G$ of $K=8$ episodes with simulated labels $\mathbf y^{\wm}$ and success probabilities $p_i$, we draw hypothetical \code{} labels $\tilde y_i^{(m)}\sim\operatorname{Bernoulli}(p_i)$ independently and compute
\begin{equation}
q(G)=\frac{1}{MK}\sum_{m=1}^{M}\sum_{i=1}^{K}\left|A_i(\tilde{\mathbf y}^{(m)})-A_i(\mathbf y^{\wm})\right|,
\label{eq:query}
\end{equation}
with task advantages $A_i$ (Appendix Eq.~\ref{eq:credit}) and $M=128$. $q(G)$ estimates the expected absolute advantage change a check would cause under the predictions, so it targets this change, not label disagreement (Figure~\ref{fig:acquisition}). Ranking by mean predicted error $\frac{1}{K}\sum_i|p_i-y_i^{\wm}|$ is \emph{error-based querying}.

\begin{table}[t]
\centering\small\setlength{\tabcolsep}{3pt}
\begin{tabular}{llrrrrr}
\toprule
& & \multicolumn{3}{c}{Primary200} & Dev500 & \\
\cmidrule(lr){3-5}\cmidrule(lr){6-6}
Reward & Query & 90 & 105 & 120 & 120 & Labels\\\midrule
Binary & Random & 50.5 & 50.5 & 53.5 & 55.0 & 0\\
Binary & Reward std. & 48.0 & 57.0 & 56.5 & 58.6 & 0\\
Binary & Constant $p$ & -- & -- & 55.5 & 54.8 & 13,248\\
Binary & Error-based & -- & -- & 59.5 & 58.2 & 13,248\\
Binary & Adv.-change & 58.5 & 55.5 & \textbf{61.0} & \textbf{62.0} & 13,248\\
WMRL & Random & 51.0 & 52.5 & 54.5 & 60.4 & 0\\
WMRL & Adv.-change & 55.0 & 56.5 & \textbf{59.5} & \textbf{62.8} & 13,248\\
\bottomrule
\end{tabular}
\caption{Query criteria after WM75 (seed 0), checking four of 16 groups per update: Primary200 strict success (\%) at updates \mbox{90--120} and Dev500 at update 120. Labels: one-time replay labels used to fit the predictor; Constant~$p$ gives every episode the labels' success rate (.514); -- marks checkpoints not evaluated. Bold marks the best endpoint for each reward processing.}
\label{tab:queries}
\end{table}

\stitle{Success probabilities.}
The score needs only a probability $p_i$ that episode $i$'s actions succeed in \code{}. Following Finding~5, we use a cheap predictor, a logistic regression over at most 30,000 TF-IDF unigram and bigram features \citep{wang2012baselines}. By default, it is fitted once on 13,248 replay labels from two earlier runs, on 711 training tasks outside all evaluation sets (Appendix~\ref{app:predictor}), then frozen and reused across learners and models. Its input is the episode as seen in \wm{}, without hidden goals or executable outcomes. Scoring 128 episodes takes about 165\,ms on one CPU thread.

\stitle{Training and cost.}
In GRPO, checked groups receive executable rewards, and the other groups keep simulated binary rewards. In WMRL \citep{yang2026wmrl}, unchecked groups keep continuous simulator scores, which Eq.~\ref{eq:query} uses in place of $\mathbf y^{\wm}$, and checked groups serve as anchors for calibration and weighting (Figure~\ref{fig:creditmechanism}). In each comparison of Appendix Table~\ref{tab:cost}, query rules check the same number of groups, and their online replay differs by at most 20\%.

\section{Does Training Benefit?}
\label{sec:results}
\stitle{Protocol.}
Runs start from the instruction model (\emph{from scratch}) or WM75 and check four of 16 groups per update. By default, policies use Qwen2.5-7B-Instruct with training seed 0 and stop at update 120, fixed in advance (Appendix Table~\ref{tab:config}); from scratch, random, error-based, and advantage-change querying also use seeds 1--2 (Appendix Table~\ref{tab:extensionseeds}). We score policies on Primary200 and a 500-task development set (Dev500) and reserve 500 more tasks (Final500) for one comparison.

\begin{figure}[t]
\centering
\includegraphics[width=\linewidth]{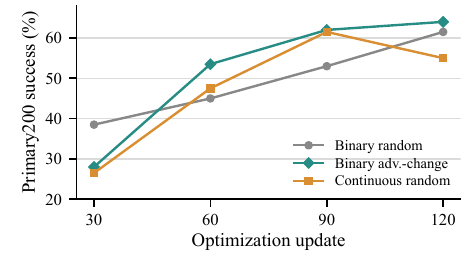}
\caption{Primary200 learning curves from scratch. Update 120 is the endpoint fixed in advance. Continuous random applies random checks to continuous simulated rewards. Table~\ref{tab:queries} gives the checkpoints after WM75.}
\label{fig:curves}
\end{figure}

\subsection{Against random and heuristic checks}
\label{sec:random}
Starting from WM75, advantage-change querying reaches 61.0\% on Primary200 and 62.0\% on Dev500 (Table~\ref{tab:queries}). Random checks reach 53.5\% and 55.0\%, checking the groups with the largest reward spread reaches 56.5\% and 58.6\%, and \mbox{error-based} querying reaches 59.5\% and 58.2\%. Replacing the learned probabilities with one constant (.514) lowers advantage-change querying to 55.5\% and 54.8\%. Offline, the constant captures nearly as much advantage change (Finding~5); we think it falls behind in training because it cannot tell which successes are false (Finding~4). Advantage-change querying also leads random checks at every evaluated update. On Dev500, it gains 64 tasks and loses 29 against random checks (exact McNemar $p<.001$; Appendix~\ref{app:cost}). Under WMRL, advantage-change anchors reach 59.5\% and 62.8\%, against 54.5\% and 60.4\% for random anchors.

\looseness=-1 From scratch, advantage-change querying beats random checks and error-based querying on both evaluation sets in all three seeds (Appendix Table~\ref{tab:extensionseeds}; Figure~\ref{fig:curves} shows seed 0). On average, it gains 2.5 and 3.0 points over random checks and 3.0 and 4.9 over error-based querying, and at least 1.4 in every seed. The default 13,248 labels cost at most 99,197 replayed actions, about three times what one from-scratch run replays online, but the predictor is trained once and reused. A 1,024-label predictor costs at most 7,686 actions, about a fifth of that, and reaches 64.5\% and 67.2\% in seed 0, against 61.5\% and 62.0\% for random checks and 60.0\% and 64.2\% for error-based querying with it (Appendix Tables~\ref{tab:scratchfull} and~\ref{tab:labelcost}).

\stitle{Other rules and held-out tasks.}
Two other rules end below random checks from scratch: ranking by predictive entropy with the same predictor (60.0\% and 57.2\%), and a reproduced CARE query-and-filter variant that trains on only 7,040 of 15,360 trajectories (44.0\% and 47.0\%; Appendix~\ref{app:extensiongrids}). The comparison fixed in advance for Final500 uses WMRL from scratch, where advantage-change anchors reach 66.4\% against 64.4\% for random anchors (332 against 322 successes). Its 95\% interval is $[-1.4,5.4]$ when paired by task and $[-3.72,8.89]$ when clustered by product (McNemar $p=.302$; Appendix Table~\ref{tab:final}). Clustering widens the interval because one product contributes 14 net successes over 17 tasks.

\stitle{Predictor choice.}
\label{sec:lightweight}
Querying with the cheap classifier also trains a better policy than querying with a more accurate neural predictor. With the same start, rule, budget, and 2,048 labels, the TF-IDF classifier reaches 58.5\% on Primary200 and 61.4\% on Dev500, and a GPU-based neural predictor that is 17 points more accurate reaches 56.0\% and 54.6\% (Appendix~\ref{app:predictor}). Accuracy alone is not enough.

\stitle{Budget and training state.}
With WM75 restarted at budgets of 12.5\%, 25\%, and 50\% (Appendix Table~\ref{tab:extensionbudgetgrpo}), advantage-change anchors beat random anchors in WMRL at every budget, by \mbox{1.5--5.0} points on Primary200 and \mbox{1.4--2.4} on Dev500. With binary rewards, advantage-change querying leads random checks at 25\% and 50\% but trails them at 12.5\% (59.5\% against 62.0\%), and error-based querying leads at 50\%. From update 120 of the from-scratch random-check run (Appendix Table~\ref{tab:laterstate}), 30 more updates with advantage-change querying reach 63.0\% and 64.4\%, against 58.5\% and 59.8\% when random checks continue.

\subsection{Reward processing}
\label{sec:conditions}
\looseness=-1 The learner's reward processing shapes what a check changes. WMRL calibrates unchecked rewards with a monotone map fitted to checked anchors and weights advantages by the anchors' residuals (Figure~\ref{fig:creditmechanism}). Calibration never reverses an ordering, but it can merge rewards into ties or change their spacing, and so change the advantages. In a WMRL run, it changes the advantages of 77 of 240 unchecked groups, including 31 without new ties (Appendix~\ref{app:processing}). Our score computes advantage changes before this processing and assumes that they approximate the changes after it. With calibration and weighting disabled at a 25\% budget, unchecked groups keep raw simulated scores. Advantage-change querying then reaches 65.0\% on Primary200 and 63.0\% on Dev500, against 47.0\% and 45.0\% for random checks. Relative to full WMRL processing, the gap over random checks grows from 5.0 to 18.0 points on Primary200 and from 2.4 to 18.0 on Dev500. At a 50\% budget and update 105, disabling both steps removes the gain (0.0 and $-2.4$ points; Appendix Table~\ref{tab:queryprocessing}).

\subsection{What changes in the policy}
Error-based querying shares our predictor, initialization, budget, and training protocol, so only the selection rule differs. In the two Dev500 tasks of Figure~\ref{fig:cases}, both from-scratch policies buy the same product but select different options. In the replay of Finding~3, the advantage-change policy's rollouts also contain fewer simulator errors than those of the random-check and error-based policies (77 against 141 and 104 per 1,024 rollouts), including fewer false successes (65 against 111 and 79), which carry most advantage error (Finding~4). For the from-scratch WMRL pair, advantage-change anchors also cut non-purchases from 70 to 6 on Dev500 and from 65 to 6 on Final500, and strict successes rise from 320 to 326 and from 322 to 332. Most added purchases miss a requirement, so finishing a purchase and meeting every requirement are separate skills (Appendix~\ref{app:policycases}).

\section{Where Does It Transfer?}
\label{sec:transfer}
\begin{table}[t]
\centering\small\setlength{\tabcolsep}{8pt}
\begin{tabular}{lrrrr}
\toprule
& \multicolumn{2}{c}{Primary200} & \multicolumn{2}{c}{Dev500}\\
\cmidrule(lr){2-3}\cmidrule(lr){4-5}
Setting & Rand. & Adv. & Rand. & Adv.\\\midrule
\multicolumn{5}{l}{\textit{Qwen2.5-7B from scratch (120 updates)}}\\
GRPO & 61.5 & \textbf{64.0} & 62.0 & \textbf{65.4}\\
WMRL & 61.0 & \textbf{61.5} & 64.0 & \textbf{65.2}\\
Dr.~GRPO & 58.0 & \textbf{62.5} & 58.0 & \textbf{62.6}\\
SDAR & 60.5 & \textbf{62.5} & 60.4 & \textbf{62.0}\\
GiGPO & 57.5 & \textbf{63.5} & 60.2 & \textbf{63.0}\\\midrule
\multicolumn{5}{l}{\textit{Qwen2.5-7B from GRPO WM75 (45 updates)}}\\
GRPO & 53.5 & \textbf{61.0} & 55.0 & \textbf{62.0}\\
WMRL & 54.5 & \textbf{59.5} & 60.4 & \textbf{62.8}\\
GiGPO & 59.5 & \textbf{60.0} & 59.4 & \textbf{61.4}\\\midrule
\multicolumn{5}{l}{\textit{Another model, original interface}}\\
Qwen3-8B & 67.0 & \textbf{69.0} & 68.0 & \textbf{68.8}\\
\bottomrule
\end{tabular}
\caption{Learners and models: strict success (\%) at update 120 with 25\% querying. Adv.: advantage-change querying with the shared 13,248-label predictor; Rand.: random querying. WMRL and Dr.~GRPO are adapted to our agent implementation; SDAR keeps its official loss and self-teacher. All rows use seed 0. Bold marks the better rule in each pair.}
\label{tab:learnertransfer}
\end{table}

\stitle{Other learners.}
The same frozen WebShop predictor supplies queries to WMRL, Dr.~GRPO \citep{liu2025understanding}, self-distilled agentic RL (SDAR; \citealp{lu2026sdar}), and group-in-group policy optimization (GiGPO; \citealp{feng2025gigpo}) (Table~\ref{tab:learnertransfer}). From scratch, advantage-change querying improves on random checks for all four other learners, by \mbox{0.5--6.0} points on Primary200 and \mbox{1.2--4.6} on Dev500. With Dr.~GRPO, advantage-change querying gains 73 Dev500 tasks and loses 50 against random checks (exact McNemar $p=.047$). After WM75, the gains for GRPO, WMRL, and GiGPO are 7.5, 5.0, and 0.5 points on Primary200 and 7.0, 2.4, and 2.0 on Dev500. With GiGPO, error-based querying is on par with advantage-change querying from both starting points (Appendix Table~\ref{tab:extensionmodels}); GiGPO also assigns step-level credit, which our task-level score does not model.

\stitle{Other models and environments.}
With Qwen3-8B \citep{yang2025qwen3} and the same predictor, advantage-change querying reaches 69.0\% on Primary200 and 68.8\% on Dev500, against 67.0\% and 68.0\% for random checks (Table~\ref{tab:learnertransfer}). These runs use seed 0, where error-based querying reaches 68.5\% and 67.8\%. In seed 1, the random and advantage-change runs collapse under our original interface, which requires a thinking tag that Qwen3's non-thinking template omits. The error-based run does not, and an action-only evaluation prompt partly restores the other two (Appendix~\ref{app:qwen}).

On frozen ALFWorld rollouts \citep{shridhar2021alfworld}, the WebShop predictor is near chance in balanced accuracy, yet its selections capture .419 mean advantage change, consistent with Finding~5. This matches a local predictor's .417 and exceeds error-based (.322) and random selection (.233; Appendix~\ref{app:transferenv}). Most ALFWorld errors are false failures (Finding~4), so this selection works under an error composition different from WebShop's. In ScienceWorld, only three of 64 groups change advantages; advantage-change and error-based selection both capture .142 there, against an expected .035 for random selection (Appendix~\ref{app:transferenv}).

\section{Conclusion}
Executable feedback and world-model training are complementary. In WebShop, an executable warm-up followed by mixing ends above either source alone. Replay shows that simulator errors are common, nearly always shift their group's advantages, and are mostly false successes. Which checks capture these shifts depends more on a group's reward structure than on predictor accuracy. Advantage-change querying turns these findings into a check rule scored on one CPU thread. In training, it outperforms random, heuristic, and error-based checks and improves other learners, and its selections transfer to frozen ALFWorld rollouts. World-model evaluations should therefore report which errors reach the learner.

\section*{Limitations}
The scope of this study follows from deliberate design choices. Checks target the reward channel, because a group-relative learner uses a label only through the advantages it sets. A check therefore replays an episode's recorded actions and replaces its label while keeping the simulated observations; combining checks with the history treatment of Finding~2, which routes executable feedback into the world model's input, is a direct way to also correct how observation errors steer later actions. We train with checks in WebShop and study further environments on frozen rollouts. Because the score needs only a group's simulated labels and a success predictor, extending training to new environments requires no change to the method.

The score is likewise learner-agnostic by design. It samples each episode's outcome independently and computes task advantages before learner-specific processing, such as WMRL's calibration, Dr.~GRPO's unnormalized advantages, or GiGPO's step-level credit, so one frozen CPU predictor serves all five learners in Table~\ref{tab:learnertransfer}. Because Eq.~\ref{eq:query} is a Monte Carlo estimate, a learner's processing and correlated outcomes can be built directly into its samples. More broadly, scoring feedback by the advantage it would change offers a way to allocate verification wherever group-relative learners train on imperfect rewards, such as LLM judges, learned reward models, or unit tests.

\section*{Ethical Considerations}
The experiments use a benchmark catalog and execute no real purchases. No human-subject data were collected. Policy evaluation uses strict task success; partial progress and purchase completion are reported as separate behavioral measurements.

\bibliographystyle{plainnat}
\bibliography{references}
\clearpage
\appendix
\section{Experimental Setup}
\label{app:setup}
This appendix gives our protocol and data. Our core code is in the supplementary software.

\begin{table}[!htbp]
\centering\small
\begin{tabularx}{\linewidth}{lrX}
\toprule
Partition & Tasks & Use\\\midrule
Training & 3,929 & Policy optimization\\
Primary & 200 & Main evaluation\\
Development & 500 & Development evaluation and comparative analysis\\
Final & 500 & One frozen comparison\\
\bottomrule
\end{tabularx}
\caption{Task partitions from a local catalog of 6,910 goals over 1,000 products. The two 500-task sets exclude training and earlier evaluation tasks by identity, goal content, and instruction. The other 1,781 goals are unused, including five that overlap other partitions.}
\label{tab:data}
\end{table}
\begin{table}[!htbp]
\centering\small\setlength{\tabcolsep}{3pt}
\begin{tabular}{ll}\toprule
Setting & Value\\\midrule
Actor & Qwen2.5-7B-Instruct\\
World model & WebShop Qwen2.5-7B\\
Learning rate & $10^{-6}$\\
KL coefficient & .01\\
Groups per update / group size & 16 / 8\\
Maximum episode actions & 15\\
Actor sampling temperature & 1.0\\
Evaluation temperature / top-p & .4 / 1\\
Evaluation seed & Fixed per task and step\\
Evaluation response tokens (max.) & 512\\
PPO minibatch / actor microbatch & 64 / 8\\
Reference log-prob microbatch & 16\\
Training seed & 0\\
Default verification budget & 4/16 groups\\
Warm-start parent / endpoint & WM75 / 120\\
From-scratch endpoint & 120\\
Monte Carlo draws per group & 128\\
\bottomrule\end{tabular}
\caption{Shared WebShop training configuration. Warm-start branches restore the complete parent, including optimizer and random-number-generator states.}
\label{tab:config}
\end{table}

Table~\ref{tab:data} lists the task partitions, and Table~\ref{tab:config} the shared configuration. Primary200 follows the public world-model setup \citep{li2026word}, and Dev500 and Final500 are further evaluation sets from the same environment. Strict success requires a purchase with an official reward of one. The official partial score reflects product type, attributes, options, and price. Buying a product other than the goal item is not a failure by itself, because other products can satisfy the instruction.

\subsection{Task advantage and checks}
\label{app:credit}
\paragraph{Environment and simulator.}
We use the 1,000-product WebShop setup released with the public text-world-model study \citep{yao2022webshop,li2026word}. The agent is Qwen2.5-7B-Instruct \citep{qwen2025qwen25}, and the simulator is the frozen \texttt{WorldModel-Webshop-Qwen2.5-7B}, which predicts the next page from the interaction history. An episode ends with a purchase or after 15 search, click, and option actions.

\paragraph{Task advantage.}
GRPO \citep{shao2024deepseekmath} in \texttt{verl-agent} \citep{feng2025gigpo} samples 16 groups of $K=8$ trajectories per update. Each trajectory receives a strict-success outcome $r_i\in\{0,1\}$, scaled to a task reward of 10 or 0. Before group normalization, each action turn with an invalid format loses 0.1. The task reward enters the update only through its group-standardized value, which we call \emph{task advantage}. For a group with outcomes $\mathbf r$ and action lengths $\mathbf L$, let $\mu_{\mathbf L}$ and $\sigma_{\mathbf L}$ be the mean and sample standard deviation after repeating each reward $L_i$ times, as in the implementation. The task advantage of trajectory $i$ is
\begin{equation}
A_i(\mathbf r,\mathbf L)=\frac{10 r_i-\mu_{\mathbf L}(10\mathbf r)}{\sigma_{\mathbf L}(10\mathbf r)+10^{-6}}.
\label{eq:credit}
\end{equation}
This is the advantage GRPO assigns from task rewards alone, excluding format penalties, masks, and gradients. A group with identical rewards has zero task advantage. A group's \emph{advantage error} is the mean absolute difference between its task advantages under simulated and executable labels.

\paragraph{Checks.}
A check resets \code{} to the same task and executes the episode's recorded actions (\emph{fixed-action replay}). An action that is invalid in \code{}, such as a click on a missing option, leaves the page unchanged, and replay moves on to the next recorded action. Replay stops when \code{} ends the episode; only an episode that ends with full reward is a success. The executable outcome replaces the simulated reward, while the simulated observations stay in the training data. A budget fixes the fraction of groups checked per update (25\%, four of 16, unless stated otherwise).

\paragraph{Runs and evaluation.}
Warm-start runs restore the complete checkpoint after 75 updates of pure world-model training (WM75), and from-scratch runs start from the instruction model. Both stop at update 120, fixed in advance.

\section{Measurement Details}
Table~\ref{tab:samples} lists the trajectory samples behind the measurements. The nine-checkpoint cohort draws new rollouts on Dev500 tasks; the historical samples reuse archived training rollouts (Appendix~\ref{app:predictor}).

\begin{table*}[t]
\centering\small\setlength{\tabcolsep}{4pt}
\begin{tabularx}{\textwidth}{ll>{\raggedright\arraybackslash}Xl}
\toprule
Sample & Size & Origin & Used in\\\midrule
Nine-checkpoint replay cohort & 1,152 groups (9,216 traj.) & Nine fixed checkpoints, 128 Dev500 tasks & Findings 3--5; Tables~\ref{tab:frozenlocalization}--\ref{tab:structural}\\
Predictor fitting pool & 13,248 labels & Two earlier runs from WM75, 711 training tasks & Default TF-IDF predictor\\
Historical validation groups & 171 groups & Same runs, 148 of 177 held-out training tasks & Finding 5; Table~\ref{tab:structural}\\
Held-out acquisition sample & 150 groups & Subset of the 171 covered by two predictors (Appendix~\ref{app:predictor}) & Figure~\ref{fig:acquisition}\\
Predictor comparison sample & 64 groups (512 traj.) & Subset of the 171 (61 tasks; 55 groups also in the 150) & Figure~\ref{fig:predictorcapacity}; Table~\ref{tab:predictor}\\
Reward-channel audit & 64 groups (512 traj.) & Latest complete groups of a pure \wm{} run, updates 1--75 & Appendix~\ref{app:channels}\\
Other text environments & 64 groups each & Frozen policies, released world models; 63 groups in TextWorld & Table~\ref{tab:alfpredictor}\\
AWM cohort & 144 groups (1,152 traj.) & Three Qwen3-8B checkpoints & Table~\ref{tab:awmcredit}\\
\bottomrule
\end{tabularx}
\caption{Trajectory samples used in the measurements. The historical samples come from training tasks and share no task with the nine-checkpoint cohort or any evaluation set (Appendix~\ref{app:predictor}).}
\label{tab:samples}
\end{table*}
\subsection{Feedback schedules and the reward channel}
\label{app:channels}
\label{app:schedules}
In the schedules of Table~\ref{tab:schedulesmain}, a group trained with full \code{} interaction receives executable observations and rewards. Checks instead only replay a simulated episode's recorded actions to obtain its executable reward, and its simulated observations stay in training.

\paragraph{Mixed \code{} schedules.}
The increasing, decreasing, and fixed-50\% schedules (the last is \emph{Random 50\% after the prefix} in Table~\ref{tab:schedulesmain}) all restore the same complete Code30 checkpoint, including the policy, optimizer, and random-number states, and continue with training seed 0 to update 150. Updates 1--30 therefore use pure \code{} in all three schedules. Each subsequent update contains 16 task groups with eight trajectories per group. Before collecting trajectories, we assign each entire group to either \code{} or \wm{}. All eight trajectories then use that environment for the full episode, obtaining both its observations and its rewards. Each group is rolled out once, without first generating and discarding a \wm{} rollout.

For cumulative update $t\in\{31,\ldots,150\}$, define the linear continuation fraction $p_t=(t-31)/119$. The numbers of \code{} groups are
\begin{align*}
k_t^{\mathrm{inc}} &= \operatorname{round}(16p_t),\\
k_t^{\mathrm{dec}} &= 16-k_t^{\mathrm{inc}},\\
k_t^{\mathrm{fixed}} &= 8.
\end{align*}
The remaining $16-k_t$ groups use \wm{}, and Table~\ref{tab:codeschedules} gives representative updates and cumulative counts. Thus the realized per-update \code{} fraction is $k_t/16$, changing in increments of $1/16$ rather than continuously. At each update, a seed-0 random permutation of the 16 group indices selects the first $k_t$ groups for \code{}. The mixing generator advances once per continuation update, and resumption restores its position in this sequence.

\begin{table}[!htb]
\centering\small\setlength{\tabcolsep}{5pt}
\begin{tabular}{lrrr}
\toprule
Update & Increasing & Decreasing & Fixed 50\%\\\midrule
1--30 & 16 & 16 & 16\\
31 & 0 & 16 & 8\\
60 & 4 & 12 & 8\\
90 & 8 & 8 & 8\\
120 & 12 & 4 & 8\\
150 & 16 & 0 & 8\\\midrule
Total, 31--150 & 960 & 960 & 960\\
Total, 1--150 & 1,440 & 1,440 & 1,440\\
\bottomrule\end{tabular}
\caption{\code{} task-group counts in the three schedules. Each update has 16 groups of eight trajectories. All schedules share the pure-\code{} prefix. The increasing schedule starts at 0\% \code{}, the decreasing schedule at 100\%, and the fixed schedule at 50\%.}
\label{tab:codeschedules}
\end{table}

\paragraph{Matched group budget.}
Each schedule uses exactly 960 \code{} groups and 960 \wm{} groups over updates 31--150. Including the 480 \code{} groups in the shared prefix, \code{} supplies 1,440 of the 2,400 groups (60\%) through update 150, corresponding to 11,520 \code{} trajectories. This budget is matched among these three Code30 continuations. The random-50\% run trained from the start instead uses 1,200 \code{} groups over 150 updates, and pure \code{} uses 2,400. Pure \code{} training reaches the same 1,440 groups at update 90, where it scores 66.0\% (Table~\ref{tab:schedulesmain}).
 
\paragraph{History treatment.}
The attribution experiments also execute each action the agent takes in \code{}. The world model still generates the observation the agent receives. Before its next prediction, however, the simulated feedback in its input context, such as observations and available actions, is replaced with the \code{} feedback. The world model therefore predicts from real history while the agent trains on simulated observations, and the reward source is set separately in each branch. The parent checkpoint of Table~\ref{tab:attributionmain}a was itself trained this way. The fixed-action checks of advantage-change querying do not use this treatment.

Table~\ref{tab:channels} asks which part of executable feedback matters when a random half of the groups receives it. Full interaction reaches 63.5\% at update 150. Applying only the history treatment to the selected groups ends at 45.0\%, below pure \wm{} training (56.5\%, Table~\ref{tab:schedulesmain}). Adding the executable terminal reward lifts the result to 55.0\%, and the terminal reward alone reaches 54.0\%. Rewards therefore matter more than the world model's history, and the remaining gap to full interaction (55.0\% against 63.5\%) reflects the observations the agent itself sees, which only full interaction makes executable. From a checkpoint trained with the history treatment only, at update 150, replacing simulated rewards with executable ones for 30 more updates reaches 60.0\%, against 51.0\% when simulated rewards are kept (Table~\ref{tab:attributionmain}a).

\begin{table}[t]
\centering\small\setlength{\tabcolsep}{4pt}
\resizebox{\linewidth}{!}{\begin{tabular}{lrrr}
\toprule
Code use in selected 50\% & 90 & 120 & 150\\\midrule
Full interaction & 54.5 & 59.0 & 63.5\\
History treatment only & 55.5 & 50.0 & 45.0\\
Terminal reward only & 52.0 & 56.5 & 54.0\\
History treatment and terminal reward & 50.5 & 59.0 & 55.0\\\midrule
\multicolumn{4}{l}{\textit{Same update-150 parent (history treatment only); update 180}}\\
Keep WM rewards & \multicolumn{3}{r}{51.0}\\
Replace with Code rewards & \multicolumn{3}{r}{60.0}\\
\bottomrule
\end{tabular}}
\caption{Which channel of executable feedback matters: Primary200 strict success (\%). Upper rows share WM75 and the same random half-group selection. Lower rows branch from one checkpoint and apply the history treatment in both branches.}
\label{tab:channels}
\end{table}

Visible discrepancies are a poor proxy for reward errors. We audit the 64 latest complete groups (512 trajectories) archived by a pure world-model run over updates 1--75. Replaying their recorded actions reaches \code{} states whose clickable options differ from the simulated ones in 456 trajectories, yet only 11 terminal labels change (seven false successes, four false failures), all within these 456. A check therefore targets terminal rewards rather than visible discrepancies. This 2.1\% error rate cannot be compared directly with the \mbox{4.4--14.0\%} of Finding~3. The audit uses archived rollouts on training tasks, and 194 of them have no terminal simulator score and count as failures. Finding~3 instead samples new rollouts from fixed checkpoints on Dev500 tasks.

\subsection{Replay cohort and error localization}
\label{app:measurementprotocol}
The nine checkpoints are fixed before collecting trajectories: pure world-model training at updates 30, 60, and 75; random checks at updates 60, 90, and 120; advantage-change checks at 60 and 120; and error-based checks at 120. Each checkpoint samples eight attempts on each of the same 128 Dev500 tasks (1,024 trajectories) with temperature 1 and sampling seed 1008, and \code{} replays every recorded action sequence. No optimizer runs, and no Final500 task is used. Within each group, we compare simulated binary rewards with executable ones and with two oracle corrections, fixing only false successes or only false failures, while actions and lengths stay fixed. Table~\ref{tab:frozenlocalization} reports the resulting reduction in advantage error. At matched updates, the rollouts of policies trained with advantage-change checks contain the fewest errors: 81 and 77 at updates 60 and 120, against 106 and 141 for random checks and 104 for error-based checks at update 120, largely because they contain fewer false successes.

\begin{table}[!htbp]
\centering\small\setlength{\tabcolsep}{3pt}
\begin{tabular}{rrrrrr}\toprule
Policy/update & FP & FN & Adv. error & Fix FP & Fix FN\\\midrule
WM/30 &37&8&.1602&79.9\%&14.6\%\\
WM/60 &118&11&.3170&88.9\%&10.2\%\\
WM/75 &131&12&.3368&89.5\%&9.2\%\\
Random/60 &79&27&.2819&72.2\%&25.4\%\\
Random/90 &90&22&.2491&84.5\%&13.8\%\\
Random/120 &111&30&.3322&81.7\%&15.8\%\\
Adv.-change/60 &62&19&.2441&74.2\%&23.9\%\\
Adv.-change/120 &65&12&.2010&83.0\%&13.2\%\\
Error/120 &79&25&.2833&79.0\%&21.0\%\\
\bottomrule\end{tabular}
\caption{Error localization on frozen rollouts. FP: false successes; FN: false failures, among 1,024 trajectories per checkpoint. The last two columns give the relative reduction in mean absolute advantage error when each oracle correction is applied alone.}
\label{tab:frozenlocalization}
\end{table}

\begin{table}[!htbp]
\centering\small
\begin{tabular}{lrr}\toprule
Checkpoint & With errors & Changed $A$\\\midrule
WM/30 & 29 & 29\\
WM/60 & 53 & 52\\
WM/75 & 56 & 54\\
Random/60 & 54 & 52\\
Random/90 & 50 & 46\\
Random/120 & 57 & 56\\
Adv.-change/60 & 43 & 43\\
Adv.-change/120 & 40 & 38\\
Error/120 & 52 & 52\\
\bottomrule\end{tabular}
\caption{Wrong-label groups per checkpoint (of 128), and how many change task advantages.}
\label{tab:errorcounts}
\end{table}

Across all 1,152 groups, 434 contain a label error and 422 of these change their task advantages (Table~\ref{tab:errorcounts}). Their median advantage error is .678 (interquartile range [.623,.876]). The 12 unchanged groups have constant labels in both sources. At update 60 of pure world-model training, task 4700 has eight false successes. At update 75, tasks 233 and 1894 show the same pattern, and their 16 wrong labels change no advantage. The reverse occurs at update 90 of random checking: task 853 has eight simulated failures that all succeed. Counting errors therefore misranks how much a group's advantages move. One wrong label in task 3184 moves its advantages by .823 on average (Appendix~\ref{app:querycases}), while the eight wrong labels of task 233 move none. Including format penalties in the advantages changes the false-success-only reductions for pure \wm{} training by less than one point.

\subsection{Structural acquisition controls}
\label{app:prevalencecontrols}
For each budget we randomly partition groups into batches of 16, rank groups within each batch, and average the realized advantage change of the selected groups over 200 partitions. Intervals resample tasks, with groups that share a task resampled together (1,000 replicates). With executable labels in hand, ranking by error count captures 89.0, 94.5, and 99.2\% of what ranking by true advantage change captures at budgets of 2, 4, and 8 of 16 groups. Table~\ref{tab:structural} compares rules that must choose before replay. The historical cohort comprises all 171 complete validation groups with TF-IDF predictions (Appendix~\ref{app:predictor}). The constant probability is the historical training success rate, $6805/13248=.514$; shuffled probabilities keep the empirical distribution of predicted probabilities but break their link to trajectories.

\begin{table}[!htbp]
\centering\small\setlength{\tabcolsep}{2.5pt}
\begin{tabular}{lrrrr}\toprule
Rule & Hist. & 95\% CI & Frozen & 95\% CI\\\midrule
Random & .235 & \na & \na & \na\\
Reward std.\ dev. & .446 & \na & \na & \na\\
Constant prob. & .420 & [.311,.534] & .406 & [.339,.475]\\
Shuffled prob. & .422 & [.304,.540] & .393 & [.327,.465]\\
Predicted error & .429 & [.315,.535] & .403 & [.320,.484]\\
Adv.-change & .471 & [.350,.582] & .444 & [.355,.539]\\
\bottomrule\end{tabular}
\caption{Captured advantage change of selection rules at a 4-of-16 budget, with task-bootstrap 95\% intervals. Hist.: 171 historical validation groups with recorded predictions; Frozen: the nine-checkpoint cohort (1,152 groups). Learned probabilities come from the 13,248-label TF-IDF classifier. \na: not computed.}
\label{tab:structural}
\end{table}

Learned probabilities also lead at a 2-of-16 budget on both cohorts, but at 8 of 16 on the frozen cohort they tie with the constant probability (.373). In training from WM75, however, the constant probability reaches 55.5\% and 54.8\%, against 61.0\% and 62.0\% with learned probabilities (Table~\ref{tab:queries}). The 128-draw estimator stays close to exact enumeration of all 256 label vectors. On the frozen cohort, with batch partitions drawn separately from those of Table~\ref{tab:structural}, the mean absolute score error is .0200, mean top-four overlap is 93.9\%, and captured advantage change is .4449 against .4454.

\subsection{What the score selects}
On the 150 held-out groups of Table~\ref{tab:samples}, both rules use the same frozen CPU predictor and one shuffle into batches of 16. Cumulative rounding selects 38 groups per rule at a 25\% budget, and 30 selections are shared (Figure~\ref{fig:acquisition}). The groups chosen only by the advantage-change score capture a mean realized advantage change of .526, against .316 for those chosen only by error. Five of the eight error-only groups are labeled all-success by the simulator. When a check confirms or uniformly flips such a group, its rewards stay constant and its advantages do not move. Seven of the eight advantage-only groups have mixed simulated outcomes. The advantage-only groups are also longer (10.27 against 5.80 actions per episode) and cost more to replay (upper bounds of 82.1 against 46.4 recorded actions per group); over all 38 selected groups, the bounds are 67.4 and 59.9.

\subsection{Do label disagreement and advantage error differ beyond WebShop?}
\label{app:awm}
\begin{table}[t]
\centering\small\setlength{\tabcolsep}{3pt}
\begin{tabular}{lrr}
\toprule
Reference vs.\ proxy & Count & \%\\\midrule
Different labels (trajectories) & 1,147/1,152 & 99.57\\
Identical task advantage (groups) & 139/144 & 96.53\\
\bottomrule
\end{tabular}
\caption{Code reference versus LLM-judge rewards on the same AWM cohort. Groups with identical task advantages have zero advantages under both channels.}
\label{tab:awmcredit}
\end{table}

\begin{table*}[t]
\centering\small\setlength{\tabcolsep}{5pt}
\begin{tabular}{lrrrrrr}
\toprule
& \multicolumn{2}{c}{ALFWorld} & \multicolumn{2}{c}{ScienceWorld} & \multicolumn{2}{c}{TextWorld}\\
Held-out metric & Transfer & Local fit & Transfer & Local fit & Transfer & Local fit\\\midrule
Accuracy (\%) & 87.50 & 90.63 & 97.66 & 97.66 & 81.55 & 86.31\\
Balanced accuracy (\%) & 49.89 & 66.00 & 50.00 & 50.00 & 50.00 & 63.32\\
Brier score $\downarrow$ & .200 & .067 & .144 & .017 & .171 & .091\\
Random: advantage change & .233 & .233 & .000 & .000 & \textemdash & \textemdash\\
Error: advantage change & .322 & .398 & .142 & .142 & \textemdash & \textemdash\\
Adv.-change: advantage change & .419 & .417 & .142 & .142 & \textemdash & \textemdash\\
Random: Code actions & 5,695 & 5,695 & 5,149 & 5,149 & \textemdash & \textemdash\\
Error: Code actions & 5,072 & 4,566 & 4,492 & 4,501 & \textemdash & \textemdash\\
Adv.-change: Code actions & 4,970 & 4,689 & 4,500 & 4,501 & \textemdash & \textemdash\\
\bottomrule\end{tabular}
\caption{Frozen-predictor diagnostics in other environments: 2,048 fitting and 512 task-disjoint validation trajectories per environment (504 in 63 complete groups for TextWorld). In ALFWorld and ScienceWorld, query rules select 16 of 64 validation groups; TextWorld reports predictor metrics only. Transfer uses the 13,248 WebShop labels and local fitting each environment's 2,048 labels. Random rows report one fixed-seed draw of four groups per batch of 16.}
\label{tab:alfpredictor}
\end{table*}

Agent World Model (AWM) supplies generated tool environments backed by databases \citep{wang2026awm}. On three fixed tasks, we sample 16 groups of eight trajectories from each of three Qwen3-8B checkpoints, a source policy at update 20 and its continuations with proxy and code-reference rewards at update 60 (1,152 episodes in 144 groups). We score the saved trajectories with the environment's code reference and with a fixed Qwen3-8B SQL-assisted judge. Here task advantages use equal trajectory weights and the population standard deviation, $A_i=(r_i-\bar r)/(s_r+10^{-6})$. The two channels disagree on almost every label, yet nearly every group has identical (zero) task advantages under both, because each channel gives constant rewards within those groups (Table~\ref{tab:awmcredit}). Label disagreement and advantage error thus differ in a second environment as well. In one task, the judge accepts registrations that conflict with an existing booking, which the code reference correctly rejects; in another, the code reference demands a state change although the goal already holds, so neither channel is uniformly preferable across tasks.

\subsection{Does the score transfer to other environments?}
\label{app:transferenv}
For ALFWorld \citep{shridhar2021alfworld}, a frozen policy at update 40 generates eight world-model attempts on each of 320 training games. Of these, 256 games (2,048 labels) fit a local predictor and 64 disjoint games (512 labels) validate it. The cohort contains 45 false successes and 97 false failures, the reverse of WebShop's error composition. ScienceWorld and TextWorld \citep{wang2022scienceworld,cote2018textworld} use the same protocol with their released world models. In Table~\ref{tab:alfpredictor}, the transferred WebShop predictor has higher Brier scores and near-chance balanced accuracy in all three environments, yet in ALFWorld its advantage-change selections capture as much advantage change as those of the local predictor. In ALFWorld, advantage-change selection is almost unaffected by which predictor supplies the probabilities (.419 and .417), whereas error-based selection drops from .398 with the local predictor to .322 with the transferred one. In ScienceWorld, three of 64 validation groups change advantages. Advantage-change and error-based selection both capture .142 there. A single fixed-seed random draw misses all three groups and captures .000, against an expected .035 over draws. TextWorld drops one group whose world-model response hit the 8,192-token limit.

\section{Method Details}
\subsection{Success predictor}
\label{app:predictor}
The CPU predictor uses TF-IDF word unigrams and bigrams with at most 30,000 features, minimum document frequency three, and sublinear term frequency, followed by logistic regression ($C=1$, at most 500 iterations). Its input is the instruction and the episode as seen in \wm{}; executable replay supplies the binary labels. On one CPU thread (macOS ARM64), a 128-episode batch needs a median of 133.3\,ms for prediction and 31.4\,ms for Eq.~\ref{eq:query} over 20 repetitions, and cold loading takes 2.33\,s. The neural alternative is the public InnerJev model \citep{innerjev2026} with its vision tower removed (4.2B parameters), used zero-shot or adapted with LoRA (rank 64) or full language-model tuning for 32 updates. Table~\ref{tab:predictor} compares the seven predictors offline, where the default classifier is 82.2\% accurate.

\paragraph{Fitting and validation data.} The predictor is fitted and validated on separate training tasks. Its labels come from two earlier runs that continued WM75: their simulated episodes were archived during training and later replayed in \code{} with the same actions. A random split by task divides their 888 tasks: 711 supply the 13,248 fitting labels, and the other 177 supply 3,312 validation trajectories. These contain 171 complete eight-trajectory groups on 148 tasks, the historical validation groups of Finding~5. The 150-group sample of Figure~\ref{fig:acquisition} keeps the groups whose eight trajectories all have predictions from both the TF-IDF classifier and a LoRA-tuned Qwen2.5-7B success classifier, which skips inputs over 8,192 tokens; the 64-group sample of Table~\ref{tab:predictor} is another subset of the 171, sharing 55 groups with it. None of these tasks appears in Primary200, Dev500, or Final500, but all sets use the same product catalog. The archive does not record each episode's update, so we do not assign the labels to one checkpoint. The 1,024- and 2,048-label pools are nested random subsets of the 13,248 fitting labels, and the TF-IDF and InnerJev predictors share them.

\begin{table}[t]
\centering\small\setlength{\tabcolsep}{3pt}
\begin{tabular}{lrrrr}
\toprule
Predictor & Labels & Acc. & Brier$\downarrow$ & $\Delta A$\\\midrule
TF-IDF logistic & 2,048 & 73.63 & .171 & .431\\
TF-IDF logistic & 13,248 & 82.23 & .145 & .360\\
InnerJev zero-shot & 0 & 69.34 & .208 & .355\\
InnerJev LoRA & 1,024 & 90.43 & .092 & .176\\
InnerJev LoRA & 2,048 & 90.23 & .081 & .410\\
InnerJev full & 1,024 & 89.65 & .092 & .434\\
InnerJev full & 2,048 & 91.02 & .082 & .356\\
\bottomrule
\end{tabular}
\caption{Offline predictor comparison on 512 held-out training-task trajectories (64 groups). $\Delta A$ is the advantage change captured when each predictor selects 25\% of groups by expected advantage change.}
\label{tab:predictor}
\end{table}

In the policy substitution (Table~\ref{tab:jevpolicy}), the 2,048-label TF-IDF and fully tuned InnerJev predictors share the label subset, the complete WM75 state, the query rule, and the budget. The neural predictor is frozen during training and needs GPU inference; the CPU predictor reaches 58.5\% on Primary200 and 61.4\% on Dev500 at update 120, against 56.0\% and 54.6\%. From scratch, the CPU predictor fitted on 1,024 labels reaches 64.5\% and 67.2\% (Table~\ref{tab:scratchfull}).

\begin{table}[!htbp]
\centering\small\setlength{\tabcolsep}{4pt}
\begin{tabular}{lrrr}
\toprule
Predictor & Labels & 90 & 120\\\midrule
CPU TF-IDF logistic & 13,248 & 58.5 & 61.0\\
CPU TF-IDF logistic & 2,048 & 57.5 & 58.5\\
InnerJev full language & 2,048 & 55.5 & 56.0\\
\bottomrule\end{tabular}
\caption{Predictor substitution on Primary200 (seed 0), from WM75 with 25\% querying. The 13,248-label row is the default predictor.}
\label{tab:jevpolicy}
\end{table}

\subsection{WMRL adaptation}
\label{app:onlinecalibration}
In our WebShop adaptation of WMRL \citep{yang2026wmrl}, checked groups (anchors) receive binary \code{} rewards, and unchecked groups keep continuous \wm{} scores. Online debiasing fits isotonic regression, over 10 score bins, to paired simulated scores and executable labels, then calibrates unchecked scores before group normalization. The pairs come only from checks in the current run, one per checked trajectory, and pairs without a \wm{} score are skipped. The first fit waits for 200 pairs; later fits follow every 64 new pairs and use the latest 5,000. Until the first fit, unchecked scores stay raw. Inverse-variance denoising then upweights anchor advantages. For each nonconstant anchor group, it measures the mean squared difference between centered calibrated and executable rewards, divided by the executable reward variance. An exponential moving average $d$ of this ratio has a half-life of 64 groups. After 32 eligible groups, the scale is fixed as $c=1/d_{\mathrm{warmup}}$, and the weight is $w=\min(4,1+cd)$; before that, and for unchecked groups, $w=1$. The weights multiply normalized advantages before the actor-loss reduction, whose denominator is not renormalized by their sum. For WMRL, Eq.~\ref{eq:query} compares the sampled labels with the uncalibrated \wm{} scores in place of $\mathbf y^{\wm}$.

\section{Training and Transfer Results}
\label{app:extensiongrids}
This section reports the endpoint results behind the main-text comparisons as Primary200 (P200) and Dev500 (D500) strict success (\%).

\subsection{From scratch}
Table~\ref{tab:extensionseeds} repeats the from-scratch comparison of Section~\ref{sec:results} with three training seeds, and Table~\ref{tab:scratchfull} adds heuristic and low-label controls for seed 0. In every seed, advantage-change querying ends above random checks, by 2.0--3.0 points on P200 and 1.4--4.2 on D500, and above error-based querying, by 1.5--5.5 and 3.8--6.0 points. More labels do not consistently make querying better. From scratch, the 1,024-label predictor gives the best result (64.5\% and 67.2\%), whereas after WM75 the 13,248-label predictor beats the 2,048-label one (Table~\ref{tab:jevpolicy}). Offline, the 13,248-label classifier even captures less advantage change than the 2,048-label one (Table~\ref{tab:predictor}). Continuous simulator scores with random checks fall to 55.0\% on P200, and WMRL's calibration and weighting bring them back to 61.0\%, close to binary rewards (61.5\%).

\begin{table}[!htbp]
\centering\small\setlength{\tabcolsep}{4pt}
\begin{tabular}{lrrr}
\toprule
Seed & Random & Error & Adv.-change\\\midrule
0 & 61.5/62.0 & 62.0/60.4 & 64.0/65.4\\
1 & 61.5/61.4 & 58.0/59.6 & 63.5/65.6\\
2 & 60.5/62.4 & 62.0/60.0 & 63.5/63.8\\\midrule
Mean & 61.2/61.9 & 60.7/60.0 & 63.7/64.9\\
\bottomrule\end{tabular}
\caption{Qwen2.5-7B trained from scratch with GRPO and 25\% querying: P200/D500 strict success (\%) at update 120 for three training seeds.}
\label{tab:extensionseeds}
\end{table}

\begin{table}[!htb]
\centering\small\setlength{\tabcolsep}{3pt}
\begin{tabular}{lrrr}
\toprule
Reward / query rule & Labels & P200 & D500\\\midrule
Binary / random & 0 & 61.5 & 62.0\\
Binary / reward std.\ dev. & 0 & 60.5 & 63.2\\
Binary / adv.\ change & 13,248 & 64.0 & 65.4\\
Binary / predicted error & 13,248 & 62.0 & 60.4\\
Binary / adv.\ change & 1,024 & 64.5 & 67.2\\
Binary / predicted error & 1,024 & 60.0 & 64.2\\
Continuous WM / random & 0 & 55.0 & 60.2\\
WMRL / random anchors & 0 & 61.0 & 64.0\\
WMRL / adv.\ anchors & 13,248 & 61.5 & 65.2\\
\bottomrule\end{tabular}
\caption{From-scratch query criteria and low-label controls (seed 0, 25\% querying, update 120). Labels count the historical replay labels that fit the predictor.}
\label{tab:scratchfull}
\end{table}

\subsection{Budgets and training state}
\begin{table}[!htb]
\centering\small\setlength{\tabcolsep}{3pt}
\begin{tabular}{llr}
\toprule
Budget & Query & P200/D500\\\midrule
\multicolumn{3}{l}{\textit{Binary GRPO}}\\
12.5\% & Random & 62.0/62.4\\
12.5\% & Error & 59.5/58.8\\
12.5\% & Adv.-change & 59.5/59.6\\
25\% & Random & 53.5/55.0\\
25\% & Error & 59.5/58.2\\
25\% & Adv.-change & 61.0/62.0\\
50\% & Random & 58.5/59.4\\
50\% & Error & 62.5/62.8\\
50\% & Adv.-change & 60.5/62.6\\\midrule
\multicolumn{3}{l}{\textit{WMRL}}\\
12.5\% & Random & 59.5/58.4\\
12.5\% & Adv.-change & 61.5/59.8\\
25\% & Random & 54.5/60.4\\
25\% & Adv.-change & 59.5/62.8\\
50\% & Random & 60.0/61.8\\
50\% & Adv.-change & 61.5/63.2\\
\bottomrule\end{tabular}
\caption{Endpoints at update 120 as Primary200/Dev500 strict success (\%), restarting the same WM75 state with seed 0. The 12.5/25/50\% budgets check 90/180/360 groups.}
\label{tab:extensionbudgetgrpo}
\label{tab:extensionbudgetwmrl}
\end{table}

\begin{table}[!htb]
\centering\small\setlength{\tabcolsep}{4pt}
\begin{tabular}{lrrr}\toprule
Query & P135 & P150 & D150\\\midrule
Random & 62.0 & 58.5 & 59.8\\
Adv.-change & 68.0 & 63.0 & 64.4\\
\bottomrule\end{tabular}
\caption{Later training state: binary-reward continuations from the update-120 checkpoint of the from-scratch random-check run, 30 further updates and 120 checked groups each. P: Primary200; D: Dev500 at update 150.}
\label{tab:laterstate}
\end{table}

\begin{table*}[t]
\centering\small\setlength{\tabcolsep}{5pt}
\begin{tabular}{llrrr}
\toprule
Query budget / endpoint & Feedback processing & Random P/D & Adv.-change P/D & Query gain P/D\\\midrule
50\% / 105 & Calibration + weighting & 54.0 / 53.0 & 56.0 / 59.2 & $+2.0$ / $+6.2$\\
50\% / 105 & Both disabled & 57.5 / 61.0 & 57.5 / 58.6 & 0.0 / $-2.4$\\
25\% / 120 & Calibration + weighting & 54.5 / 60.4 & 59.5 / 62.8 & $+5.0$ / $+2.4$\\
25\% / 120 & Both disabled & 47.0 / 45.0 & 65.0 / 63.0 & $+18.0$ / $+18.0$\\
\bottomrule\end{tabular}
\caption{Comparisons with and without WMRL's calibration and weighting: Primary200/Dev500 strict success (\%). All runs restore WM75 with seed 0.}
\label{tab:queryprocessing}
\end{table*}

Table~\ref{tab:extensionbudgetgrpo} restarts the same WM75 state at three budgets for binary GRPO and WMRL. With WMRL, advantage-change anchors lead random anchors at every budget. With binary rewards, the order of the rules changes with the budget: random checks lead at 12.5\%, advantage-change querying at 25\%, and error-based querying at 50\%. The binary random-check runs alone span 53.5--62.0\% on P200 across budgets. The 50\% random run and the terminal-reward run of Table~\ref{tab:channels} both restore WM75 and give a random half of the groups executable rewards, yet they reach 58.5\% and 56.5\% at update 120. They are separate runs: the terminal-reward run picks its groups before the rollout and executes their actions in \code{} alongside \wm{}, whereas the random-check run replays recorded actions after the rollout, and the two draw different random groups. Table~\ref{tab:laterstate} continues the from-scratch random-check run from update 120 for 30 more updates. Advantage-change querying leads at both evaluated updates, by 6.0 points at update 135 and 4.5 at update 150.

\subsection{Reward processing}
\label{app:processing}
Table~\ref{tab:queryprocessing} crosses the selection rule with WMRL's calibration and weighting. Disabling calibration feeds raw scores to normalization, and disabling weighting fixes all weights to one. Turning both off removes the query gain at a 50\% budget and enlarges it at 25\%, where the disabled random control falls to 47.0\%. The blocks differ in budget and endpoint, so we compare only within them. The disabled 25\% advantage-change run reaches 65.0\% on Primary200, the highest value in the table.

Calibration also changes advantages on its own, and our score is computed before this processing. Across all 240 unchecked groups from updates 76--105 of a WMRL run with random checks on half of the groups, calibration reverses no strict ordering but merges 257 unequal pairs into ties across 46 groups. Task advantages change by more than $10^{-5}$ in 77 groups, including 31 without new ties. Even without the denominator stabilizer, 26 of these 31 changes remain, and ten occur in groups whose episodes all end normally. Table~\ref{tab:calibrationcase} shows one such group: every ordering is preserved, yet trajectory 7 moves from above to below the group mean and its advantage flips from $+0.20$ to $-1.07$ (Figure~\ref{fig:creditmechanism} illustrates the mechanism).

\begin{table}[!htb]
\centering\small\setlength{\tabcolsep}{3pt}
\begin{tabular}{rrrrrr}
\toprule
$i$ & $L_i$ & $r_i$ & $\hat r_i$ & $A_i$ & $\hat A_i$\\\midrule
1 & 5 & 0.333 & 0.000 & $-$1.5262 & $-$1.2074\\
2 & 5 & 1.000 & 0.756 & $+$0.7816 & $+$0.8669\\
3 & 13 & 0.400 & 0.031 & $-$1.2954 & $-$1.1229\\
4 & 5 & 1.000 & 0.756 & $+$0.7816 & $+$0.8669\\
5 & 5 & 1.000 & 0.756 & $+$0.7816 & $+$0.8669\\
6 & 10 & 1.000 & 0.756 & $+$0.7816 & $+$0.8669\\
7 & 5 & 0.833 & 0.048 & $+$0.2047 & $-$1.0746\\
8 & 5 & 1.000 & 0.756 & $+$0.7816 & $+$0.8669\\
\bottomrule\end{tabular}
\caption{A complete group (task 239, update 84) from a WMRL run with random checks. $L_i$: action count; $r_i,\hat r_i$: raw and calibrated rewards; $A_i,\hat A_i$: the corresponding task advantages. The ordering is unchanged, but trajectory 7 crosses zero.}
\label{tab:calibrationcase}
\end{table}

\subsection{Other learners and models}
Table~\ref{tab:extensionbaselines} adds entropy ranking with the same predictor, a reproduced CARE query-and-filter rule, and two learners trained from instruction weights; Table~\ref{tab:extensionmodels} lists the remaining learner runs. Against random checks, advantage-change querying gains 4.5 and 4.6 points with Dr.~GRPO (Dev500: 73 gains, 50 losses, exact McNemar $p=.047$) and 2.0 and 1.6 points with SDAR. SDAR's teacher is the current actor itself, which rescores the sampled responses before each update with SDAR's WebShop skill bank in its prompt. With GiGPO, error-based querying is on par with advantage-change querying from both starts (Table~\ref{tab:extensionmodels}). GiGPO's step-level credit lies outside our task-level score.

Entropy ranking scores a group by the mean Bernoulli entropy of its eight predicted success probabilities and checks the highest. It reaches 60.0\% and 57.2\%, below advantage-change querying from the same start (64.0\% and 65.4\%). The reproduced CARE rule follows CARE's query-and-filter design, with task labels and frozen CPU probe features in place of CARE's answer clusters and hidden-state features. Two heads on these features learn from every check: a reliability head predicts whether all eight labels of a group are correct, and a correction head predicts how much a check would correct the group's advantages. Like our rule, it checks four of 16 groups per update, but it also drops groups. In the first 20 updates, it checks the four groups with the largest predicted correction and trains only on them. Afterwards, it keeps the four groups most likely to be fully correct, checks the four of the remaining 12 with the largest predicted correction weighted by the chance of a wrong label, and drops the other eight. It therefore trains on 7,040 of the 15,360 generated trajectories, whereas our rule trains on all of them.

\begin{table}[!htb]
\centering\small\setlength{\tabcolsep}{3pt}
\begin{tabular}{llrr}
\toprule
Setting & Query & P200 & D500\\\midrule
Reproduced CARE & Query + filter & 44.0 & 47.0\\
Same predictor & Entropy & 60.0 & 57.2\\
Dr.~GRPO & Random & 58.0 & 58.0\\
Dr.~GRPO & Adv.-change & 62.5 & 62.6\\
SDAR & Random & 60.5 & 60.4\\
SDAR & Adv.-change & 62.5 & 62.0\\
\bottomrule\end{tabular}
\caption{Additional selection rules and learners, from scratch, Qwen2.5-7B, 25\% querying, seed 0, update 120. Dr.~GRPO is the objective adapted to our agent implementation; SDAR keeps its official loss and uses the current actor with SDAR's skill bank as teacher.}
\label{tab:extensionbaselines}
\end{table}

\begin{table}[!htb]
\centering\small\setlength{\tabcolsep}{3pt}
\begin{tabular}{llrr}
\toprule
Learner / model & Query & P200 & D500\\\midrule
GiGPO & Random & 57.5 & 60.2\\
GiGPO & Error & 62.5 & 63.6\\
GiGPO & Adv.-change & 63.5 & 63.0\\\midrule
GiGPO from WM75 & Random & 59.5 & 59.4\\
GiGPO from WM75 & Error & 60.5 & 61.2\\
GiGPO from WM75 & Adv.-change & 60.0 & 61.4\\
\bottomrule\end{tabular}
\caption{GiGPO endpoints at update 120, 25\% querying, seed 0. The last block continues the GRPO WM75 checkpoint with GiGPO.}
\label{tab:extensionmodels}
\label{tab:extensioninitialization}
\end{table}

\paragraph{Qwen3 interface.}
\label{app:qwen}
The original training interface requires an explicit thinking tag that Qwen3's non-thinking template does not produce. Under this interface, seed 1 collapses for random and advantage-change querying, to 9.0\% and 7.0\% on Primary200, while error-based querying does not (Table~\ref{tab:qwenpromptgrid}). The collapsed advantage-change policy, for example, uses up its 512-token budget without an executable action in 2,345 of 2,798 turns. Error-based querying stays between 67.6\% and 69.5\% across seeds, prompts, and evaluation sets. Evaluating the same checkpoints with an action-only prompt, without retraining, raises the collapsed policies to 44.5\% and 68.0\% but lowers the seed-0 random policy from 67.0\% to 56.0\%. With this prompt, advantage-change querying ends above random checks in both seeds. Qwen3 results are thus highly sensitive to the interface and the evaluation prompt.

\begin{table}[!htb]
\centering\small\setlength{\tabcolsep}{2.5pt}
\begin{tabular}{llrrrr}
\toprule
& & \multicolumn{2}{c}{Original} & \multicolumn{2}{c}{Action-only}\\
Seed & Query & P200 & D500 & P200 & D500\\\midrule
0 & Random & 67.0 & 68.0 & 56.0 & 59.4\\
0 & Error & 68.5 & 67.8 & 69.5 & 68.0\\
0 & Adv.-change & 69.0 & 68.8 & 70.5 & 71.0\\\midrule
1 & Random & 9.0 & 14.0 & 44.5 & 52.6\\
1 & Error & 68.0 & 67.6 & 68.5 & 68.4\\
1 & Adv.-change & 7.0 & 8.4 & 68.0 & 70.0\\
\bottomrule\end{tabular}
\caption{Qwen3-8B at update 120 under the original interface, strict success (\%) for two training seeds. Each checkpoint is evaluated with the original prompt and, without retraining, with an action-only prompt. Evaluation temperature is .4.}
\label{tab:qwenpromptgrid}
\end{table}

\section{Final500 and Execution Costs}
\label{app:cost}
The Final500 comparison was fixed before accessing the set and evaluates the from-scratch WMRL pair at update 120 (Table~\ref{tab:final}). Advantage-change anchors gain 2.0 points (exact McNemar $p=.302$). The largest-gain target product contributes 14 net successes over 17 tasks, so the product-cluster interval is wider than the paired one. On Dev500, advantage-change querying gains 64 tasks and loses 29 against random checks after WM75 (exact McNemar $p<.001$). Table~\ref{tab:cost} lists online execution costs, which differ by less than a fifth within each comparison. Historical labels are a one-time cost (Table~\ref{tab:labelcost}). The 1,024-label pool costs at most 7,686 replayed actions, about a fifth of the 35,629 actions that random checks replay online when training from scratch.

\begin{table}[!htb]
\centering\small\setlength{\tabcolsep}{5pt}
\begin{tabular}{lr}
\toprule
Final500 statistic & Value\\\midrule
Successes: random / Adv.-change & 322 / 332\\
Gains / losses & 43 / 33\\
Difference (percentage points) & $+2.0$\\
Paired 95\% interval & $[-1.4,5.4]$\\
Product-cluster 95\% interval & $[-3.72,8.89]$\\
Exact McNemar $p$ & .302\\
Target-product clusters & 140\\
\bottomrule\end{tabular}
\caption{The comparison fixed in advance on Final500: from-scratch WMRL, random versus advantage-change querying, update 120, one training seed.}
\label{tab:final}
\end{table}

\begin{table}[!htb]
\centering\small\setlength{\tabcolsep}{3pt}
\begin{tabular}{llr}
\toprule
Setting & Query & Actions\\\midrule
WM75, binary & Random & 11,455\\
WM75, binary & Adv.-change & 11,671\\
WM75, WMRL & Random & 9,958\\
WM75, WMRL & Adv.-change & 11,413\\
WM75, WMRL disabled & Random & 9,783\\
WM75, WMRL disabled & Adv.-change & 10,548\\
Scratch, binary & Random & 35,629\\
Scratch, binary & Error & 33,137\\
Scratch, binary & Adv.-change & 35,985\\
Scratch, WMRL & Random & 35,247\\
Scratch, WMRL & Adv.-change & 30,102\\
\bottomrule\end{tabular}
\caption{Online \code{} actions replayed for checks up to update 120 (25\% querying). Warm-start counts cover the 45 updates after WM75 and from-scratch counts all 120; historical labels are a one-time cost (Table~\ref{tab:labelcost}).}
\label{tab:cost}
\end{table}

\begin{table}[!htb]
\centering\small
\begin{tabular}{rr}
\toprule
Historical labels & Recorded-action upper bound\\\midrule
1,024 & 7,686\\
2,048 & 15,425\\
13,248 & 99,197\\
\bottomrule\end{tabular}
\caption{One-time cost of historical labels, bounded above by recorded action lengths.}
\label{tab:labelcost}
\end{table}

\section{Behavioral Analysis and Cases}
\label{app:cases}
We examine concrete actions behind the measurements, query decisions, and policy comparisons. We chose the examples to illustrate these behaviors; they do not estimate how often each behavior occurs. The measurement examples use recorded simulated rollouts and executable replays; the policy examples are evaluated entirely in \code{}.

\subsection{What a simulator error changes for the learner}
\label{app:measurementcases}
\paragraph{Every success label is wrong, yet the task advantages stay zero (task 233).}
The task asks for gray, medium-size women's shorts with elastic closure and faux fur, costing less than \$60. In one rollout of the frozen WM75 policy, the agent searches for these requirements, opens a pair of bike shorts, selects \texttt{gray} and \texttt{medium}, and clicks \texttt{buy now}. The world model returns a purchase confirmation with score 1.0. Replaying those same actions in \code{} purchases the same item with the same options but returns 0.0. Even the search results differ.

Across all eight rollouts for this task, \wm{} reports eight successes and \code{} reports eight failures. Nevertheless, both give zero task advantage to every attempt. The simulator says that every attempt did equally well, and execution says that every attempt did equally badly, so neither distinguishes an attempt to favor over the others. Correcting all eight labels therefore leaves the group-relative task signal unchanged.

\paragraph{Mixing supplies an option that both pure policies omit (task 3561).}
The request is for a quick-release, hands-free phone accessory for wireless charging, with color option \texttt{windshield kit} and style \texttt{windshield}, under \$70. We compare the update-150 policies in Table~\ref{tab:schedulesmain}: pure \code{}, pure \wm{}, and random 50\% full interaction after Code30 (all seed 0). All three find the same product. Their final actions differ:
\begin{quote}\small
Pure \code{} and pure \wm{}: select \texttt{windshield kit}, then buy.\\
Mixed: select \texttt{windshield}, select \texttt{windshield kit}, then buy.
\end{quote}
Both pure policies leave the style unset and receive $5/6$ rather than full success. The mixed policy sets both options and succeeds. Here the gain comes from completing the purchase configuration, with the product held fixed.

\paragraph{Finding the item and finishing the purchase are different skills (task 653).}
Under the same update-150 comparison, the task requires a machine-washable shirt with specified fabric attributes, cranberry color, women's fit, and medium size, under \$50. The pure-\code{} policy repeatedly returns to search and opens alternative products. It reaches the eventual successful product on action 14 and selects the fit on action 15, exhausting its allowance before purchase. The pure-\wm{} policy reaches that product on action 2 and has selected the color, fit, and size by action 5. It then reselects these options until action 15 and never buys.

The mixed policy reaches the product on action 2, selects women's fit, cranberry, and medium, and buys on action 6 for full success. Thus the two pure policies fail at different points in this example: one spends its actions searching, and the other keeps manipulating an already configured item. The mixed policy completes both parts.

\subsection{Which groups advantage-change querying brings to a check}
\label{app:querycases}
The next examples come from the 171 historical validation groups used to evaluate the frozen CPU selector. Each contains eight simulated attempts with executable replay labels. We rank this held-out pool by predicted advantage change and inspect groups within the top 25\% (43 groups). The displayed scores were computed from simulated trajectories before consulting their replay labels.

\paragraph{A false success encourages buying the wrong item (task 3184).}
The task requests a 3.5-ounce product for meat or seafood with the listed ingredient requirements, under \$20. Several attempts open a 3.5-ounce meat rub, select its size, and buy successfully. One attempt initially opens that item but returns to search and eventually buys a children's protein shake. The simulated page for the shake lists 8- and 16-fluid-ounce packs and a price of \$100. The agent nevertheless issues \texttt{click[3.5 ounce]} twice and buys. The world model calls this a success, whereas executable replay rejects it.

The selector ranks this group eighth of 171, with predicted advantage change 1.123. Replay changes its success pattern from seven successes and one failure to six successes and two failures. The protein-shake attempt moves from positive task advantage ($+0.236$) to negative advantage ($-1.638$): it should be discouraged, rather than rewarded alongside the successful purchases. The mean absolute advantage change is 0.823 even though just one of the eight labels changes. This group shows why even one error can make a check useful.

\paragraph{Correct purchases should not be discouraged (task 1623).}
The request specifies a machine-washable tuxedo-themed shirt, heather grey, men's fit, size 3t, under \$40. One attempt opens the matching shirt, selects \texttt{men}, \texttt{heather grey}, and \texttt{3t}, then buys. The world model assigns only 0.6, which becomes a failure under strict success, while the same actions achieve full success in \code{}. Three other attempts complete the requirements but are also labeled failures by \wm{}. One genuinely unsuccessful attempt buys a different shirt and never selects the requested color.

The selector ranks this group fifth, with predicted advantage change 1.159. Replay increases the number of successes from three to seven. The four wrongly rejected attempts switch from negative to positive task advantage, while the remaining failure stays negative. The measured mean absolute change is 1.196. Checking this group therefore restores credit to the correct options and purchases.

The same task is selected again online after WM75, at update 84, and replaying its new attempts turns four failures into successes, a mean advantage change of 1.291.

\subsection{What the learned policy does better than random querying}
\label{app:policycases}
These Dev500 examples compare random and advantage-change querying from the instruction model, with binary rewards, a 25\% budget, seed 0, and update 120, both evaluated in \code{}.

\paragraph{Replace a redundant click with the missing size (task 3860).}
The request is for workout shorts with a drawstring closure and elastic waist, color \texttt{\#1 grey navy}, size 30, under \$40. Both policies find the same shorts and finish in five actions. After opening the product, their actions are:
\begin{quote}\small
Random: \texttt{click[\#1 grey navy]}, repeat the color click, \texttt{click[buy now]}.\\
Advantage-change: \texttt{click[\#1 grey navy]}, \texttt{click[30]}, \texttt{click[buy now]}.
\end{quote}
Random leaves the size unset and scores $5/6$. Advantage-change supplies the missing size and scores 1.0. The improvement is visible in a single action, with the same product and action budget.

\paragraph{Keep the product that satisfies both attributes (task 1817).}
The user wants high-definition binoculars for bird watching, under \$870. Both policies first open the same suitable binoculars. Random then returns to search, changes its query to \texttt{binoculars for bird watching price<870.00}, and buys another product. That product satisfies one of the two requested attributes and scores $2/3$. The advantage-change policy buys the first product in three actions and satisfies both attributes for score 1.0. The successful product differs from the task's goal item, showing that success here comes from meeting the requirements rather than reproducing a particular product identifier. The error-based policy also buys an item satisfying just one attribute in this task.

\paragraph{Finish after finding and configuring the right item (task 2606).}
The task asks for easy-clean bathing accessories for removing dead skin, in the \texttt{rose red 1 pair} option, under \$60. Random opens the appropriate scrubbing mitts on action 5 and selects the requested option on actions 6 and 7. It then returns to search, visits other products, and reaches the 15-action limit without buying. The advantage-change policy opens the mitts on action 2, selects the same option twice, and buys on action 5 for full success. Both trajectories repeat an option click, but only one completes the purchase.

\paragraph{The paired examples in Figure~\ref{fig:cases}.}
Task 3192 (apricot eye makeup) compares the same advantage-change policy with error-based querying. Both buy the same brush set. Error-based querying buys after 12 actions without choosing a color, whereas advantage-change selects apricot and buys after five, raising the score from .8 to 1. Its color click changes the option state without changing the visible page. The reverse occurs in task 382: only the error-based policy selects the required size, and advantage-change leaves it unset. These cases show both a recovered constraint and a remaining failure.

\paragraph{Purchase completion across tasks.}
For the separate from-scratch WMRL pair, advantage-change querying reduces non-purchases from 70 to 6 on Dev500 and from 65 to 6 on Final500, and raises strict successes from 320 to 326 and from 322 to 332. On Dev500, 55 of the 70 former non-purchases become failed purchases, and failed purchases rise from 110 to 168 (113 to 162 on Final500). Completing a purchase and satisfying every requirement remain distinct outcomes, as the action-level examples illustrate.

\end{document}